%% file: acl2022.tex
\pdfoutput=1

\documentclass[11pt]{article}

\usepackage[]{acl}

\usepackage[T1]{fontenc}

\usepackage[utf8]{inputenc}

\usepackage{hyperref}
\usepackage{microtype}
\usepackage{times}
\usepackage{latexsym}
\usepackage[ruled,linesnumbered]{algorithm2e}
\usepackage{amssymb}
\usepackage{amsfonts}
\usepackage{amsmath} 
\usepackage{amsthm} 
\usepackage{booktabs}
\usepackage{enumerate}
\usepackage{graphicx}
\usepackage{subfigure}
\usepackage{xspace}
\usepackage{float}
\usepackage{bbm}
\usepackage{bm}
\usepackage{multirow}
\usepackage{booktabs}
\usepackage{color}
\usepackage{framed}
\usepackage{stfloats}
\usepackage{iitem}
\usepackage{makecell}
\usepackage[normalem]{ulem}
\useunder{\uline}{\ul}{}

\usepackage{url}

\newcommand{\ie}{\emph{i.e.,}\xspace}

\newcommand{\eg}{\emph{e.g.,}\xspace}

\newcommand{\ignore}[1]{}

\usepackage{xcolor}
\definecolor{tOrange}{RGB}{255,165,0}
\definecolor{tBlue}{RGB}{24,116,205}
\definecolor{tPink}{RGB}{255,20,147}
\definecolor{tGreen}{RGB}{50,205,50}
\definecolor{tGold}{RGB}{255,215,0}

\title{\includegraphics[width=1.5em, trim=25 25 25 25]{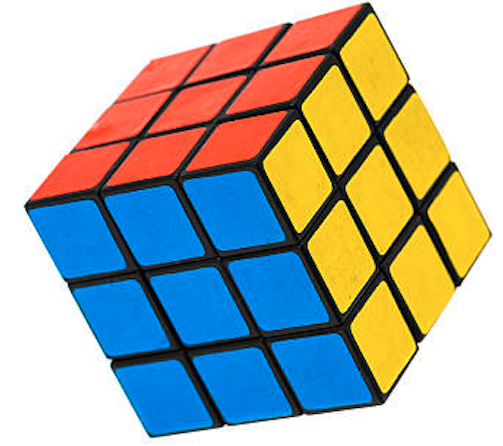} \textsc{Data-CUBE}: Data Curriculum for Instruction-based Sentence Representation Learning}

\author{
        Yingqian Min \textsuperscript{\rm{1}}\footnotemark[1],
        Kun Zhou \textsuperscript{\rm{1}}\footnotemark[1],
        Dawei Gao \textsuperscript{\rm{3}},
	Wayne Xin Zhao\textsuperscript{\rm{2}}\footnotemark[2] , 
	He Hu\textsuperscript{\rm{1}}, 
	{\rm and} Yaliang Li\textsuperscript{\rm{3}} \\
 \textsuperscript{1}School of Information, Renmin University of China \\
	\textsuperscript{2}Gaoling School of Artificial Intelligence, Renmin University of China \\
	\textsuperscript{3}Alibaba Group \\
	\texttt{\{yingqianm,hehu\}@ruc.edu.cn},  \texttt{francis\_kun\_zhou@163.com} \\
        \texttt{batmanfly@gmail.com}, \texttt{\{gaodawei.gdw,yaliang.li\}@alibaba-inc.com} \\
}

\begin{document}
\maketitle
\footnotetext[1]{Equal contribution.}
\footnotetext[2]{Corresponding author.}

\begin{abstract}
Recently, multi-task instruction tuning has been applied into sentence representation learning, which endows the capability of generating specific representations with the guidance of task instruction, exhibiting strong generalization ability on new tasks.
However, these methods mostly neglect the potential interference problems across different tasks and instances, which may affect the training and convergence of the model.
To address it, we propose a data curriculum method, namely Data-CUBE, that arranges the orders of all the multi-task data for training, to minimize the interference risks from the two views.
In the task level, we aim to find the optimal task order to minimize the total cross-task interference risk, which is exactly the traveling salesman problem, hence we utilize a simulated annealing algorithm to find its solution.
In the instance level, we measure the difficulty of all instances per task, then divide them into the easy-to-difficult mini-batches for training.
Experiments on MTEB sentence representation evaluation tasks show that our approach can boost the performance of state-of-the-art methods. Our code and data are publicly available at the link: \url{https://github.com/RUCAIBox/Data-CUBE}.


\end{abstract}

\input{sec-intro}

\input{sec-rel}

\input{sec-pre}

\input{sec-model}

\input{sec-exp}
\input{sec-con}



\bibliography{ref}
\bibliographystyle{acl_natbib}

\newpage

\end{document}

%% file: sec-intro.tex
\section{Introduction}
\label{sec-intro}


Sentence representation learning~\cite{Reimers-2019-sbert, Gao-2021-SimCSE} is a fundamental task in the NLP area, which focuses on encoding the semantic information of sentences into low-dimensional vectors.
Typically, existing work~\cite{Karpukhin-2020-DPR, Wang-2022-arxiv-E5} collects a set of sentence pairs (or augmented in unsupervised way), and then optimizes the model parameters to maximize and minimize the similarity of relevant and irrelevant sentences, respectively.
Previous methods~\cite{Zhou-2023-ECML-Master, Zhou-2022-DCLR} based on advanced language models and learning objectives\cite{Zhou-2022-Simans}, are capable of producing high-quality representations that perform well on various intended tasks, \eg text semantic matching~\cite{Agirre-2012-STS12, Agirre-2013-STS13} and classification~\cite{O'Neill-2021-arxiv-Iwish, Casanueva-2020-arxiv-Efficient}.

\begin{figure}[t]
    \centering
    \includegraphics[width=0.5\textwidth, trim=10 30 0 15]{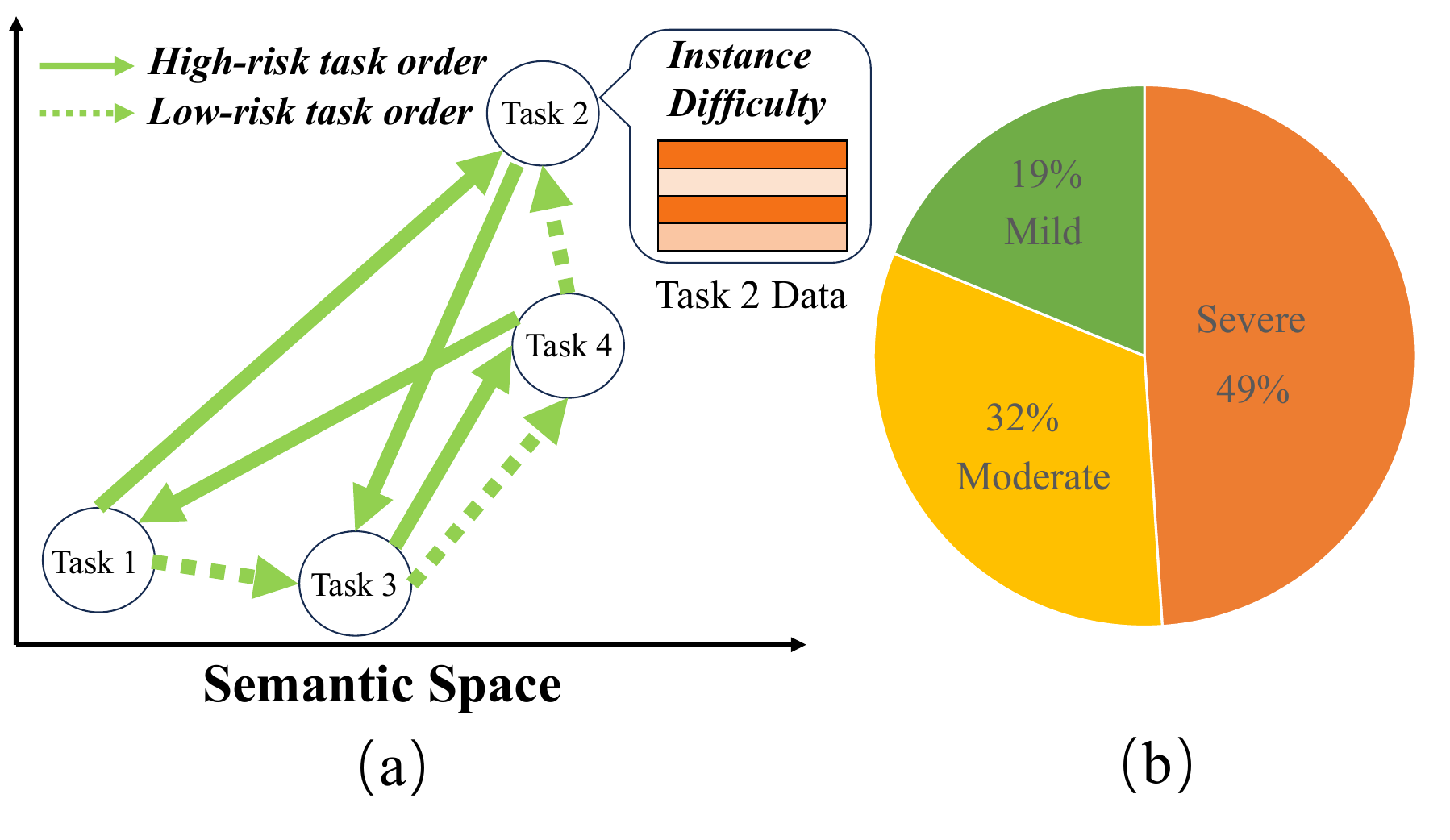}
    \caption{\textbf{(a)} \textbf{Example of the task and instance interference}. 
    The distance reflects task similarity, and the shades of oranges represent the difficulty level.
    \textbf{(b)} \textbf{The underfitting degrees of all training tasks}. We categorize all tasks into three degrees: severe~(>80\%), moderate~(>50\% but <80\%), and mild~(<50\%), according to the ratio of instances whose positives and negatives are not clearly distinguished (margin<0.05).}
    \label{fig:intro}
\end{figure}

Despite the success, recent studies~\cite{Neelakantan-2022-Text} have revealed that it is challenging to directly transfer the learned sentence representations into new tasks, even causing significant performance degradation.
To address it, instruction tuning~\cite{Wei-2022-Flan, Wang-2022-Tk_instruct} has been applied to sentence representation learning, which collects a diverse set of sentence-pair datasets with task-specific natural language instructions~\cite{Su-2023-INSTRUCTOR, Xiao-2023-BGE}.
After multi-task training on the dataset collection, the model would be capable of generating specific sentence representations with the guidance of the task instruction, exhibiting strong generalization ability on unseen tasks.


Before training, each of the collected datasets is generally divided into multiple mini-batches at random~\cite{Su-2023-INSTRUCTOR}, and the representation model will learn the mini-batches of all datasets in a random order.
However, as the collected datasets vary in data distributions, their random order would lead to potential cross-task interference risk for the diversity of tasks' optimization direction. 
As depicted in Figure~\ref{fig:intro}, once the neighboring mini-batches are from very different tasks, the successive learning of them would lead to conflict in the optimization objective, affecting the learning of both tasks.
Besides, the instances within the same task might also own different difficulty.
Randomly assigning them into mini-batches may also cause the potential cross-instance interference risk, further influencing the task learning.
Consequently, the cross-task and cross-instance interferences would impede the representation model to well fit the training data, increasing the underfitting risk.

\begin{figure*}[h]
    \centering
    \includegraphics[width=1\textwidth]{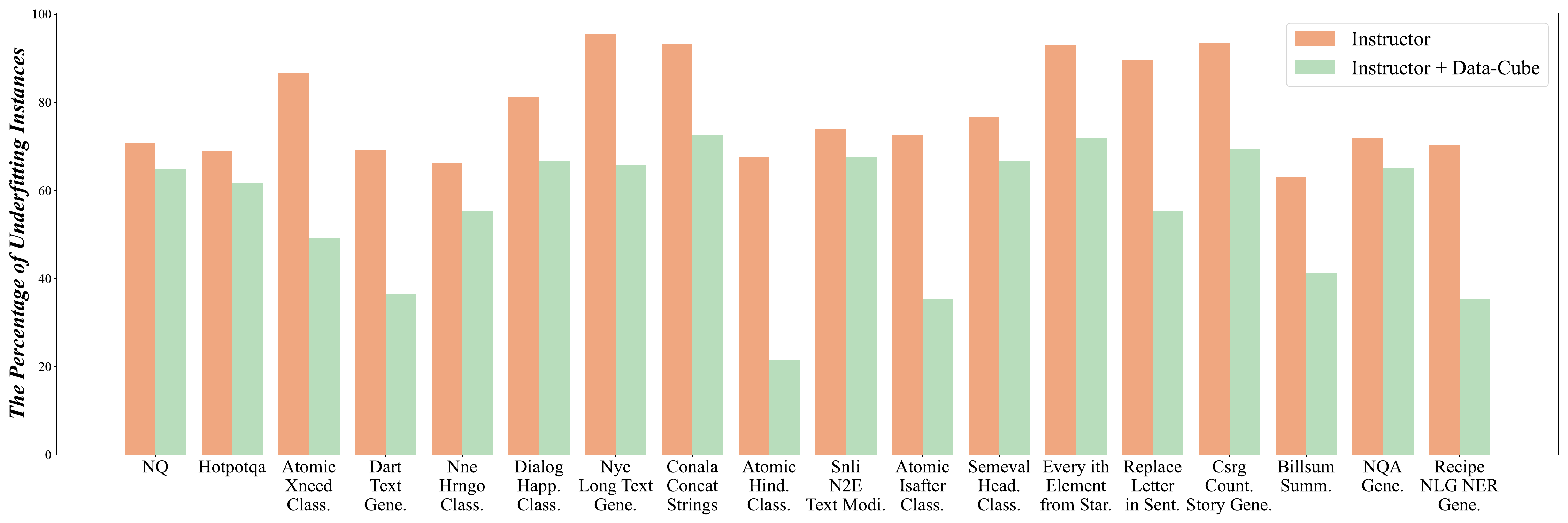}
    \caption{The proportion of underfitting instances within different tasks. We show the comparison between INSTRUCTOR and fine-tuned INSTRUCTOR with Data-CUBE.}
    \label{fig:underfitting_degree}
\end{figure*}

We conduct experiments to explore the extent of underfitting in the mainstream instruction-based sentence representation models~(\eg INSTRUCTOR~\cite{Su-2023-INSTRUCTOR}).
As shown in Figure~\ref{fig:intro}, INSTRUCTOR is struggling with distinguishing more than 80\% positive and negative examples in almost half of the training datasets, indicating a pronounced underfitting problem. To further investigate what kind of tasks suffer from the underfitting issue, we randomly sample multiple tasks from all 330 tasks and calculate the ratio of underfitting instances~(See Figure~\ref{fig:underfitting_degree}). The analysis reveals that the presence of underfitting phenomena is pervasive in various task categories, not limited to specific tasks.

To address the above problems, we aim to design a proper data curriculum that arranges the orders of all tasks and instances for minimizing the potential interference risks.
Concretely, for the cross-task interference, we focus on finding the optimal task order where all the neighboring two tasks are as similar as possible, to minimize the total interference risks derived from task divergence.
By regarding all the tasks as the nodes in a fully connected graph and computing the similarity of all tasks as the edge weights, we convert the optimal order search problem into finding the longest route that visits all nodes.
It is exactly the travelling salesman problem~(TSP)~\cite{Hoffman-2013-traveling}, hence we can use the widely-studied simulated annealing~(SA) algorithm to efficiently find its suboptimal solution.
For the cross-instance interference, we measure the discriminability of positive and negative, as the estimated difficulty for sorting all the instances.
Then, we divide them into the easy-to-difficult mini-batches for training, to minimize the interference risks caused by varied instance difficulty.

To this end, we propose \textbf{\textsc{Data-CUBE}}, a \underline{Data} \underline{CU}rriculum method for instruction-\underline{B}ased sentenc\underline{E} representation learning.
In our approach, with the help of a pre-learned representation model, we first estimate the cross-task similarity and instance difficulty.
Then, we employ the SA algorithm based on the cross-task similarity to find the optimal order for task arrangement, and sort the instances within each task according to their difficulty to obtain the easy-to-difficult order for mini-batch arrangement.
Finally, we divide all the instances from all the tasks into a sequence of mini-batches for the training of the sentence representation model.
Extensive experiments have shown that our approach can consistently improve the performance of state-of-the-art baselines, even outperforming larger models using much more data.

Our contributions are summarized as follows:

(1) To our knowledge, our approach is the first attempt of data curriculum in instruction-based sentence representation learning.

(2) We reveal the serious data interference problem in instruction-based sentence representation learning, and propose Data-CUBE, a novel method using task-level and instance-level curriculums for alleviating this problem.  Data-CUBE serves as a model and data-agnostic method. By virtue of its pre-reorganization of data, it avoids incurring any additional training costs compared to direct instruction tuning.

(3) Experimental results on 28 downstream tasks show the effectiveness of our approach. 
As Figure~\ref{fig:underfitting_degree} shows, our method significantly improves the underfitting problem across various tasks.

%% file: sec-rel.tex
\section{Related Work}
We review the related work from the following three aspects: sentence representation learning, instruction tuning, and traveling salesman problem.

\paragraph{Sentence Representation Learning.} A robust sentence representation plays a pivotal role in diverse downstream tasks. Previously, most sentence representation models concentrate on a singular task or domain, weak in transferring to other downstream tasks without further fine-tuning. For instance, SimCSE~\cite{Gao-2021-SimCSE}, SBERT~\cite{Reimers-2019-sbert}, and DCLR~\cite{Zhou-2022-DCLR} are trained to address sentence similarity and classification tasks, while models like DPR~\cite{Karpukhin-2020-DPR}, Contriever~\cite{Izacard-2022-Contriever}, Master~\cite{Zhou-2023-ECML-Master}, and GTR~\cite{Ni-2022-GTR} are applied to information retrieval. In response to this challenge, recent efforts have emerged to develop instruction-based sentence representation models through multi-task contrastive learning. Exemplars include INSTRUCTOR~\cite{Su-2023-INSTRUCTOR}, BGE~\cite{Xiao-2023-BGE}, and GTE~\cite{Li-2023-GTE}, which aim to enhance the adaptability and generalization capabilities of sentence representations across diverse tasks and domains. 
Existing studies\cite{Kun-2023-CLASSP-L2P} primarily focus on aspects such as the training objective\cite{Zhou-2022-Simans}, model architecture\cite{Zhou-2022-DCLR}, or training scale, while paying limited attention to the challenges posed by interference during the multi-task training process.

\paragraph{Instruction Tuning.} 
Instruction tuning~\cite{Ouyang-2022-InstructGPT, Zhao-2023-arxiv-LLMSurvey} involves supervised fine-tuning pre-trained language models by integrating well-formatted natural language instructions into the input. This process is closely connected to multi-task learning and is believed to enhance the generalization capability of language models across a range of tasks~\cite{Wei-2022-Flan}. Previous studies have demonstrated that increasing the number and diversity of tasks associated with instructions can improve performance is achievable through instruction tuning.

Considering the effectiveness of instruction tuning, it has been applied to various NLP tasks, such as sentence representation learning~\cite{Su-2023-INSTRUCTOR, Xiao-2023-BGE}. However, with the increasing diversity of tasks, there is a potential for interference across different tasks, which may lead to performance degradation~\cite{Mueller-2022-ACL-Do}. Hence, we propose to leverage data curriculum to alleviate the interference in the multi-task instruction tuning.

\paragraph{Traveling Salesman Problem.} \textbf{T}raveling \textbf{S}alesman \textbf{P}roblem~(TSP) is a classic combinatorial optimization problem, whose goal is to find the shortest possible route for a traveling salesman to visit a given set of cities exactly once and return to the starting city~\cite{Hoffman-2013-traveling, Cheikhrouhou-2021-TSP_survey}. The problem has been extensively studied in computer science and operations research and has been proven to be an NP-hard problem, indicating that it is generally difficult to solve in polynomial time. 
Therefore, a large number of heuristic methods~\cite{Helsgaun-2006-Effective, Matai-2010-traveling_overview} have been well studied, which can find the suboptimal solution in a small error range and with a reasonable time for millions of cities. 
Among them, \textbf{S}imulated \textbf{A}nnealing~(SA)~\cite{Bertsimas-1993-Simulated} is a simple and effective optimization algorithm to obtain an approximate solution~\cite{Aarts-1988-Quantitative, Otubamowo-2012-Comparative, Delahaye-2019-Simulated_app}. 
It is inspired by the annealing process in metallurgy, which involves heating a material to a high temperature and then gradually cooling it to remove defects. 
Simulated annealing is widely used to find approximate solutions for combinatorial optimization problems, \eg TSP, Job Shop Scheduling Problem~\cite{Chakraborty-2015-IJHIT-SA2JSH}, and Graph Coloring Problem~\cite{Pal-2012-PCS-SA2GCP}.


%% file: sec-pre.tex
\section{Preliminary}
\label{sec-pre}
In this section, we formulate the instruction-based sentence representation learning task and then introduce the traveling salesman problem. 

\subsection{Problem Statement}
Sentence representation learning is to train an encoder that can map a sentence into a latent vector for intended tasks.
However, traditional methods may suffer performance degradation when transferred to unseen tasks~\cite{Neelakantan-2022-Text}.
To address it, the recently proposed instruction-based sentence representation learning~\cite{Su-2023-INSTRUCTOR} takes the sentence $s$ with a natural language instruction $I$ as the input, to obtain the task-aware sentence representation $\textbf{v}$.
By training on multiple sentence-pair datasets with corresponding instructions, the sentence representation model can follow new instructions to flexibly adapt the representations into new tasks without any training, achieving remarkable cross-task transferring performance.

Formally, to train the instruction-based sentence representation model, we are given $m$ instruction-based sentence-pair datasets $\mathcal{D}=\{d_{i}\}_{i=1}^{m}$, where $d_i$ denotes the $i$-th dataset and corresponds to the task $o_i$.
Each dataset typically consists of $n$ queries $\{q_{j}\}_{j=1}^{n}$ and their relevant sentences $\{s^{(+)}_{j}\}_{j=1}^{n}$ and irrelevant sentences $\{s^{(-)}_{j}\}_{j=1}^{n}$, with specific instructions $\langle I^{(q)}, I^{(+)}, I^{(-)} \rangle$ for the three text types.
During training, the model follows a certain task order $\mathcal{O} = \{o_i\}$ to learn the corresponding task data.
However, due to the inconsistent data distribution and objectives, the tasks and instances may interfere with each other during training, affecting the model learning.
In this work, we devise a data curriculum approach to organizing the training order of all the multi-task data, to reduce the interference risk.

\subsection{Travelling Salesman Problem}
Considering the interference problem across the $m$ datasets, we first estimate the mutual interference risks between every two tasks $r(i,j)$, and then find the optimal order for all the tasks $\mathcal{O} = \{o_i\}_{i=1}^{m}$, to minimize the accumulated interference risk as:
\begin{equation}
   \arg\min \sum_{i=1}^{m-1} r(o_{i}, o_{i+1}) + r(o_m, o_1) \label{eq-TSP}
\end{equation}
where $o_i$ is a task dataset from $\mathcal{D}$.
Such an optimal order search problem can be converted as the traveling salesman problem~(TSP) that finds the shortest route to visit all cities (\ie tasks) exactly once.

As TSP is proved to be an NP-hard problem, heuristic algorithms~\cite{Helsgaun-2006-Effective, Matai-2010-traveling_overview} have been widely studied to find suboptimal solutions in a reasonable time.
Simulated annealing~(SA)~\cite{Bertsimas-1993-Simulated} is a commonly used algorithm for TSP, 
and its basic idea is to start with an initial solution and then search by randomly perturbing the solution. 
In each iteration, the algorithm evaluates the quality of the new solution by computing the change in the objective function. 
If the new solution is better, it will replace the current one. 
If not, the update would occur based on a probability based on the temperature and the change in the objective function.
As the iterations progress, the temperature decreases, reducing the likelihood of accepting worse solutions and guiding the algorithm towards converging.

%% file: sec-model.tex
\section{Approach}
In this section, we present our devised data curriculum method to alleviate the mutual interference issue in instruction-based sentence representation learning.
Concretely, we adopt multi-task contrastive learning to train the representation model on a set of sentence-pair datasets, and mainly focus on arranging the orders of all tasks and instances before training.
In the task level, we design a task distance-based metric to estimate the interference risk, then leverage the SA algorithm to find the suboptimal task order.
In the instance level, we mainly consider the interference derived from different instance difficulty, and rearrange the instances to follow the easy-to-difficult order.

\begin{figure*}[h]
    \centering
    \includegraphics[width=1\textwidth]{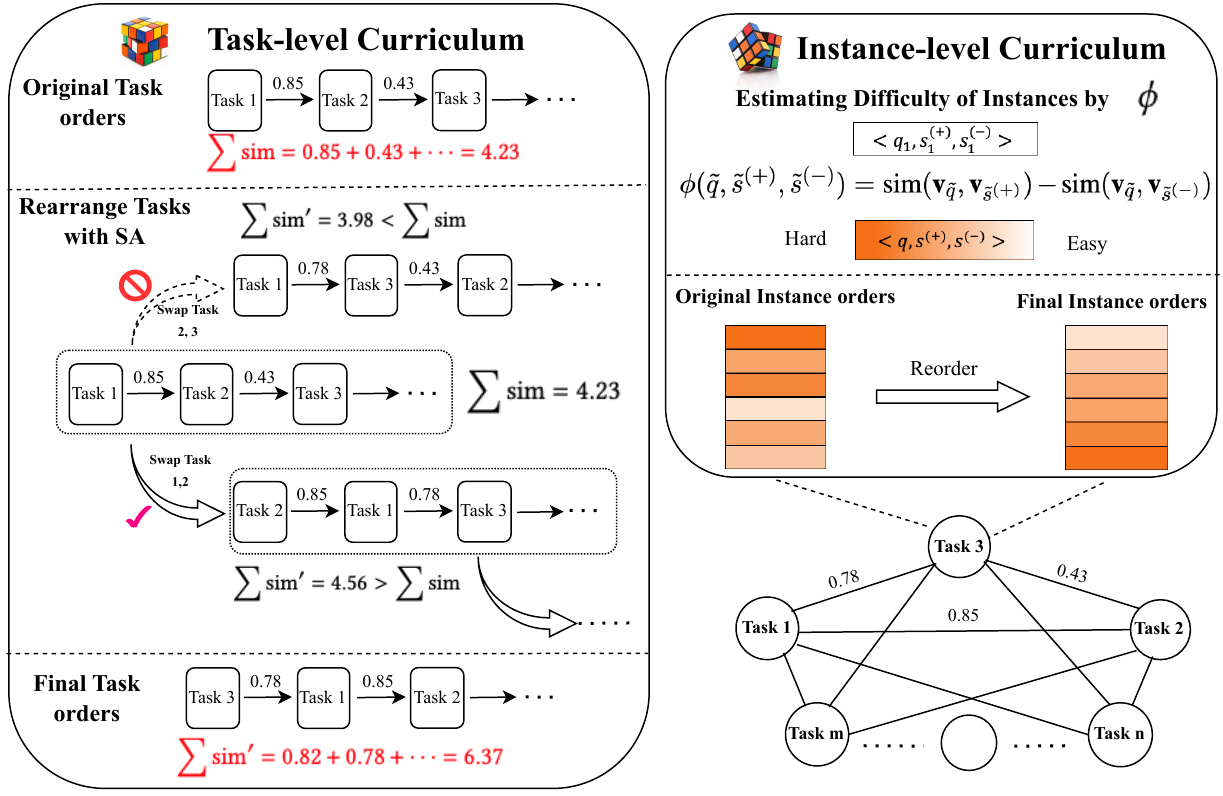}
    \caption{Illustration of Data-CUBE: Task-level Curriculum rearranges the task orders from similar to dissimilar using Simulated Annealing and Instance-level Curriculum reorganizes the instances within each task from easy to difficult.}
    \label{fig:cube}
\end{figure*}

\subsection{Multi-task Contrastive Learning}
We employ multi-task contrastive learning to train a pre-trained language model (\eg T5~\cite{Raffel-2020-JMLR-T5}) for producing sentence representations, which maximize the similarity of positive pairs $\langle q,s^{(+)} \rangle$ and minimize the one of negative pairs $\langle q,s^{(-)} \rangle$, with the consideration of specific task instructions $\langle I^{(q)}, I^{(+)}, I^{(-)} \rangle$.
Concretely, we first preprocess the instances from all the datasets into multiple mini-batches with specific instructions and then optimize the model parameters via a multi-task learning loss.

For each dataset, we concatenate its contained queries, positive and negative sentences with corresponding instructions, to compose new instances:
\begin{equation}
\small
    \Tilde{q} = [I^{(q)}; q], \Tilde{s}^{(+)} = [I^{(+)}; s^{(+)}], \Tilde{s}^{(-)} = [I^{(-)}; s^{(-)}]
\end{equation}
where the instruction contains the description that specifies the corresponding task, \eg ``\emph{Represent the example for the following task: Given a scientific question, generate a correct answer to it}''.
Next, we split the processed datasets into multiple mini-batches, and guarantee that all the in-batch instances come from the same task.
Such a way avoids the possible cross-task interference when using in-batch negatives for contrastive learning.
Thus, we leverage the following loss function:

\begin{equation}
\mathcal{L} = \sum_{i=1}^{m}\sum_{\mathcal{B}\in d_i} \sum_{j=1}^{\left|\mathcal{B}\right|}
\frac{e^{\text{sim}(\textbf{v}_{\Tilde{q}_{j}}, \textbf{v}_{\Tilde{s}_{j}}^{(+)}) / \tau}}
{\sum_{k=1}^{\left|\mathcal{B}\right|}
e^{\text{sim}(\textbf{v}_{\Tilde{q}_{j}}, \textbf{v}_{\Tilde{s}_{k}}) / \tau} } \label{eq-cl}
\end{equation}

where $\mathcal{B}=\{\langle \Tilde{q}_{j}, \Tilde{s}_{j}^{(+)}, \Tilde{s}_{j}^{(-)} \rangle\}_{j=1}^{|\mathcal{B}|}$ denotes the mini-batch from the dataset $d_i$, and $|\mathcal{B}|$ is the mini-batch size.
$\textbf{v}_{\Tilde{q}_{j}}$ and $\textbf{v}_{\Tilde{s}_{j}}^{(+)}$ refer to the representations of the $j$-th query $\Tilde{q}_{j}$ and positive sentence $\Tilde{s}_{j}^{(+)}$, respectively,
$m$ denotes the number of datasets and $\tau$ denotes the temperature,
$\text{sim}(\textbf{x}, \textbf{y})$ is the cosine similarity between two representations, \ie $\frac{\textbf{x}^{\top} \textbf{y}}{||\textbf{x}||\cdot||\textbf{y}||}$.
Based on the above loss, we propose the data curriculum method for the construction and ordering of all the mini-batches.
After training, we can adapt the sentence representation into new task by using corresponding instructions.

\subsection{Task-level Curriculum Arrangement: From Similar to Different}
\label{task_level_method}
Given the $m$ datasets for different tasks, we aim to find the most proper training order $\mathcal{O}=\{o_i\}_{i=1}^{m}$ for arranging the task-level curriculum.
As all the task data is learned by the representation model in a certain order~(\eg random), the divergence of data distributions between neighboring tasks may affect the learning of each other~\cite{Ding-2023-CVPR-Mitigating}.
Therefore, we expect that the training order can provide a smooth transition across tasks, where all the neighboring two tasks should be as similar as possible, to minimize the accumulated cross-task interference risk.
To achieve it, we first estimate the cross-task interference risk based on task similarity, and then search the optimal task order by the simulated annealing (SA) algorithm.

\subsubsection{Cross-task Interference Risk Estimation}
\label{task_level_measurement}
As all the tasks are formulated into the same format and learning objective in Eq.~\ref{eq-cl}, 
their major divergence lies in the data distribution.
Thus, we adopt the similarity of data representations to estimate the cross-task interference risk.
Concretely, we randomly sample $n_{t}$ queries per dataset to compose the representative data subset, then compute their mean representation as the task representation:
\begin{equation}
\textbf{v}^{(t)} = \frac{1}{n_{t}}\sum_{j=1}^{n_{t}}\textbf{v}_{\Tilde{q}_j}  \label{7}
\end{equation}
where we use a pre-learned model (\ie Instructor~\cite{Su-2023-INSTRUCTOR}) to produce the query representation.
Based on it, the task similarity can be measured using the cosine similarity of task representations.
As similar tasks typically have lower interference risk~\cite{Mueller-2022-ACL-Do}, we can roughly estimate the cross-task interference risk as:
\begin{equation}
    r(i,j)\propto -\text{sim}(\textbf{v}^{(t)}_i, \textbf{v}^{(t)}_j) \label{eq-rel}
\end{equation}
where $\text{sim}(\textbf{v}^{(t)}_i, \textbf{v}^{(t)}_j)$ is the cosine similarity between the representations of the $i$-th and $j$-th tasks, and $\propto$ means a direct proportion. It simply depicts the relation between cross-task interference and representation similarity, to reduce the computation cost in the following optimal order search.

\subsubsection{Optimal Order Search}
Based on the above metric, we aim to find the optimal training order $\mathcal{O}=\{o_i\}_{i=1}^{m}$ for all the $m$ task datasets, to minimize the interference risk for all the neighboring tasks.
As every two tasks can be neighboring, all the tasks can be regarded as nodes in a fully connected undirected weighted graph, where the edge weight is the interference risk between two linked tasks.
In this way, we can convert the optimal order search problem into finding the shortest route that visits each node exactly once, which is a TSP with the objective function as Eq.~\ref{eq-TSP}.

According to the negative correlation between the interference risk and the task similarity as Eq.~\ref{eq-rel}, the risk minimization objective is equivalent to maximizing the sum of neighboring task similarity:
\begin{equation}
    \arg \max \sum_{i=1}^{m} \text{sim}(\textbf{v}^{(t)}_i, \textbf{v}^{(t)}_{i+1}) + \text{sim}(\textbf{v}^{(t)}_m, \textbf{v}^{(t)}_{1}) \label{eq-sa-obj}
\end{equation}
Therefore, our goal is to find the most smooth transition path for all the tasks, to avoid the drastic distribution shift of the neighboring tasks.
To solve TSP, we adopt the SA algorithm to find a suboptimal solution within a reasonable amount of time.
The SA algorithm iteratively perturbs the current solution to explore the solution space, and accepts the new solution based on the objective function Eq.~\ref{eq-sa-obj} and a gradually decreasing temperature $\tau_{s}$.
Concretely, we first initialize a task order $\mathcal{O}^{'}$ by random shuffling.
Next, we repeat the perturb-then-check process until convergence.
In each iteration, we randomly choose a pair of tasks in the current order $\mathcal{O}^{'}$, swap their positions to obtain the new order $\mathcal{\Tilde{O}}^{'}$, and check if the total neighboring task similarity will increase.
If increased, the new order will replace the current one.
Otherwise, the new order will be accepted in a probability as:
\begin{equation}
    p(\mathcal{O}^{'}, \mathcal{\Tilde{O}}^{'}, \tau_s) = \exp({-\frac{\Delta(\mathcal{O}^{'}, \mathcal{\Tilde{O}}^{'})}{\tau_{s}}})
\end{equation}
where $\Delta(\mathcal{O}^{'}, \mathcal{\Tilde{O}}^{'})$ denotes the difference of the total neighboring task similarity between the current and new orders using Eq.~\ref{eq-sa-obj}.
Such a way prevents the solution from being stuck at a local minimum, and the usage of $\Delta$ also reduces the likelihood of accepting worse solutions.
Besides, as $\tau_{s}$ would gradually decrease to zero, it also reduces the instability close to the converged suboptimal point.

\subsection{Instance-level Curriculum Arrangement: From Easy to Difficult}
In addition to the task-level curriculum, we also devise the instance-level curriculum, to reduce the cross-instance interference risk.
For each dataset, its contained instances will also be learned by the model in a certain order, hence we aim to reduce the divergence between neighboring instances.
Here, we mainly consider the varying difficulty of instances, as existing work has shown that using instances with large difference in difficulty would affect the learning of each other~\cite{Bengio-2009-ICML-Curriculum}.
Thus, we aim to reorder the instances in each task from easy to difficult.
Specifically, we first estimate the difficulty of all instances based on the discriminability of positive and negative sentences, then sort all the instances and divide them into multiple mini-batches for training.


\subsubsection{Instance Difficulty Estimation}
\label{instance_level_measurement}
As all the given tasks focus on distinguishing the relevant sentence $\Tilde{s}^{(+)}$ and irrelevant sentence $\Tilde{s}^{(-)}$ according to the query $\Tilde{q}$, we leverage the discriminability of $\Tilde{s}^{(+)}$ and $\Tilde{s}^{(-)}$ to measure the instance difficulty.
Generally, the positive and negative sentences of easy instances would be clearly distinguished by a representation model trained on the data, while the ones of difficult instances can not.
Thus, we employ a pre-learned model (\ie Instructor) to encode the representations of the positive and negative pairs, then estimate the instance difficulty by computing the similarity difference as:
\begin{equation}
\label{eq-ins_diff}
\phi(\Tilde{q}, \Tilde{s}^{(+)}, \Tilde{s}^{(-)}) = \text{sim}(\textbf{v}_{\Tilde{q}}, \textbf{v}_{\Tilde{s}^{(+)}}) - \text{sim}(\textbf{v}_{\Tilde{q}}, \textbf{v}_{\Tilde{s}^{(-)}})
\end{equation}
The smaller the difference is, the more likely the model is to struggle with distinguishing the positive and negative, which means that the instance is difficult.

\begin{algorithm}[t]
  \small
  \SetKwInOut{Input}{Input}\SetKwInOut{Output}{Output}
  \caption{Training with Data-CUBE}
  \label{alg:sa_for_TSP}
  
  \Input{Original Data $\mathcal{D} = \left\{d_i\right\}$, the temperature $\tau$, cooling rate $\alpha$, and max iterations $N$ of SA, and the backbone model 
  } 
  \Output{Instruction-tuned model $M$'} 
    \BlankLine 
  \textbf{\textcolor[RGB]{0, 130, 0}{// Task Curriculum}} \\
    {Initialize a random order $\mathcal{O}' = \left\{o_i\right\}$} \\
    \For{$i$ in $range(N)$}{
        $\mathcal{\Tilde{O}}' \gets$ Swap a random pair of tasks in $\mathcal{O}'$ \\
        Calculate $\Delta(\mathcal{O}, \mathcal{O}')$ using Eq.~\ref{eq-sa-obj} \\
        \uIf{$\Delta(\mathcal{O}', \mathcal{\Tilde{O}}') > 0$}{
          $\mathcal{O}' \gets \mathcal{\Tilde{O}}'$
          }
        \ElseIf{$\text{rand()} < p(\mathcal{O}^{'}, \mathcal{\Tilde{O}}^{'}, \tau_s) $}{
            $\mathcal{O}' \gets \mathcal{\Tilde{O}}'$
        }
      $\tau_s \gets \alpha \cdot \tau_s$ \\
    }
    \BlankLine
   \textbf{\textcolor[RGB]{0, 130, 0}{// Instance Curriculum}} \\
       \For{$d_i$ in $\mathcal{D}$}{
            Calculate $\phi(\Tilde{q}, \Tilde{s}^{(+)}, \Tilde{s}^{(-)})$ of each instance in $d_i$ using Eq.~\ref{eq-ins_diff} \\
            Arrange $d_i$ to $d_i'$ in descending order based on $\phi$
       }
   \BlankLine
   \textbf{\textcolor[RGB]{0, 130, 0}{// Combine Two-level Curriculum}} \\
        Initialize an empty $\mathcal{D}'$ \\
       \For{$o_i$ in $\mathcal{O}$}{
            Choose dataset $d_i'$ corresponding to $o_i$ \\
            Select the first batch $\mathcal{B}$ of $d_i'$ \\
            $\mathcal{D}'.append(\mathcal{B})$ \\
            $d_i'.remove(\mathcal{B})$ \\
            
       }
    \BlankLine
\textbf{\textcolor[RGB]{0, 130, 0}{// Multi-task Contrastive training}} \\  
Use $\mathcal{D}'$ to train the backbone model with Eq.~\ref{eq-cl}\\
Get the final instruction-tuned model
  
\end{algorithm}

\subsubsection{Instance Curriculum Arrangement}
According to the above metric, we can assign estimated scores to all instances in each task.
Then, we sort all the instances in each task by descending, and divide them into multiple mini-batches $\mathcal{B}$.
As a result, we obtain the easy-to-difficult mini-batches per task.
Compared with the commonly used approach of randomly sampling the instances, our method can alleviate the interference caused by varying instance difficulty within each mini-batch and the difficulty divergence in neighboring mini-batches.

Besides, we find that too difficult instances may not always be useful, as they could potentially introduce noise into data.
Therefore, following existing work~\cite{Zhou-2022-DCLR}, we design a binary mask $\alpha_{i}$ using a threshold $\delta$ to reduce its influence as:
\begin{equation}
\label{eq-mask}
    \alpha_{i} = \begin{cases} 0, \phi(\Tilde{q}_i, \Tilde{s}_i^{(+)}, \Tilde{s}_i^{(-)}) \geq \delta \\ 1, \phi(\Tilde{q}_i, \Tilde{s}_i^{(+)}, \Tilde{s}_i^{(-)}) < \delta \end{cases}
\end{equation}
Then, we multiply the mask and the contrastive loss of each instance in Eq.~\ref{eq-cl}, which prevents the noisy instances from being learned but allows their contained sentences to be utilized as the in-batch negatives for other instances.

%% file: sec-exp.tex
\section{Experiments}


\begin{table*}[t]
    \begin{center}
    \centering
    \resizebox{1\textwidth}{!}{%
    \begin{tabular}{l|cccccccccc|c}
    \toprule
       {Model}  & {BIO} & {S-R}& {S12} & {S13} & {S14} & {S15} & {S16} &{S17} & {S22} &{S-B} & {Avg.} \\
    
    \midrule
        \midrule
        \multicolumn{12}{l}{\textbf{Sentence Representation APIs}} \\
        \midrule
        OpenAI-TE                   & 86.35   & 80.60   & 69.80  & 83.27 & 76.09 & 86.12 & 85.96 & 90.25 & 68.12 & 83.17        & 80.97    \\
        Voyage                      & 84.85   & 79.71  & 77.09 & 88.91 & 82.08 & 89.21 & 84.74 & 90.73 & 62.10  & \textbf{89.86}        & 82.93    \\
        Cohere                      & 85.01   & 82.18  & 77.62 & 85.16 & 80.02 & 88.92 & \textbf{86.92} & 90.09 & 66.81 & 88.79        & 83.15    \\
        Ember                       & 85.81   & 81.75  & 78.51 & 86.62 & 83.06 & 88.39 & 86.82 & 87.90  & 66.76 & 87.77        & 83.34    \\
    \midrule
    \midrule
     \multicolumn{12}{l}{\textbf{No-Instruction Sentence Representation Models}}\\
     \midrule
        GloVe                       & 44.93   & 55.43  & 54.64 & 69.16 & 60.81 & 72.31 & 65.34 & 77.95 & 56.35 & 61.54        & 61.85    \\
        USE                         & 78.19   & 74.43  & 72.58 & 72.22 & 69.98 & 82.22 & 76.91 & 85.22 & 61.90  & 80.28        & 75.39    \\
        Contriever                  & 83.32   & 70.20   & 64.34 & 80.03 & 74.51 & 83.30  & 79.67 & 86.32 & 64.64 & 78.81        & 76.51    \\
        GTR                         & 81.91   & 74.29  & 70.12 & 82.72 & 78.24 & 86.26 & 81.61 & 85.18 & 65.76 & 77.73        & 78.38    \\
        SimCSE                      & 68.38   & 80.77  & 75.30  & 84.67 & 80.19 & 85.40  & 80.82 & 89.44 & 61.96 & 84.25        & 79.12    \\
        SGPT                        & 79.50    & 79.59  & 74.29 & 85.35 & 79.21 & 85.52 & 82.54 & 90.44 & 63.20  & 85.67        & 80.53    \\
        E5                          & 84.73   & 80.49  & 75.93 & 85.22 & 80.54 & 88.81 & 85.28 & 89.37 & 62.99 & 87.21        & 82.06    \\
        SentenceT5                         & 80.43   & 80.47  & 78.85 & \textbf{88.94} & 84.86 & 89.32 & 84.67 & 89.46 & 65.33 & 84.01        & 82.63    \\
        \midrule
        \midrule
        \multicolumn{12}{l}{\textbf{Instruction-based Sentence Representation Models}} \\
        \midrule
        Jina                        & 84.43   & 79.2   & 74.52 & 83.16 & 78.09 & 86.91 & 83.65 & 90.16 & 64.88 & 84.60         & 80.96    \\
        Udever                      & 85.52   & 81.41  & 77.47 & 86.38 & 81.17 & 88.23 & 86.29 & 90.62 & 65.01 & 88.02        & 83.01    \\
        Stella                      & 85.94   & 81.06  & 78.72 & 84.88 & 83.11 & 88.74 & 86.35 & 87.71 & 66.28 & 87.45        & 83.02    \\
        BGE                         & 84.65   & 81.68  & \textbf{79.05} & 86.37 & 82.78 & 88.03 & 86.49 & 87.5  & 67.05 & 87.52        & 83.11    \\
        GTE                         & 88.65   & 79.81  & 76.81 & 88.11 & 82.66 & 88.93 & 84.25 & 88.47 & \textbf{69.71} & 86.07        & 83.35    \\
        

        \midrule
        INS                         & 84.39   & 81.27  & 76.28 & 88.18 & 81.92 & 89.01 & 85.49 & 90.30  & 67.74 & 86.88        & 83.15    \\
        \textbf{+Data-Cube} & \textbf{89.37} & \textbf{82.52} & {78.46} & {88.39} & \textbf{83.06} & \textbf{89.46} & {85.87} & \textbf{91.08} & {68.28} & {87.61} & \textbf{84.41} \\  
    
    \bottomrule
    \end{tabular}
    }
    \end{center}

    \caption{
        Sentence representation performance on 10 STS tasks (Spearman's correlation on the English test set) in MTEB~\cite{Muennighoff-2023-EACL-MTEB}. We choose diverse models as baselines, including traditional no-instruction sentence representation models, instruction-based sentence representation models, and sentence representation APIs. In the case of models with multiple versions (\eg varying parameter scales), we opt for the version that demonstrates superior performance. All reported results are derived from the MTEB Leaderboard. 
        We employ bold numbers to emphasize the best results obtained on each dataset. 
    }
    \label{tab:main_sts}
    \vspace{-5pt}
\end{table*}

\begin{table}[t]
    \begin{center}
    \centering
    \small
    \begin{tabular}{c|lcc}
    \toprule
       {Task Type} &{Datasets} & {INS} & \textbf{{+Data-CUBE}}   \\
    
    \midrule

    \midrule
        \multirow{5}{*}{Reranking}&AUDQ &64.30 & \textbf{64.74}  \\
        &SODQ &52.17 & 51.96  \\
        &SDRR &82.00 & \textbf{82.82}  \\
        &MSR &31.68 & \textbf{31.73}  \\
        \cmidrule(lr){2-4}
        &Avg. &57.53 & \textbf{57.81}  \\
    \midrule
        \multirow{12}{*}{Clustering}&ACP2P & 43.16 & \textbf{43.76} \\  
        &ACS2S  & 32.56 & \textbf{33.25} \\
        &BCP2P  & 37.62 & \textbf{37.63} \\
        &BCS2S  & 31.33 & {31.06} \\
        &MCP2P  & 34.22 & {33.98} \\
        &MCS2S  & 32.00 & {30.89} \\
        &RC     & 64.65 & {63.56} \\
        &RCP2P  & 64.63 & \textbf{65.31} \\
        &SEC    & 68.78 & \textbf{70.23} \\
        &SECP2P & 36.15 & {35.59} \\
        &TNC    & 54.13 & \textbf{55.82} \\
        \cmidrule(lr){2-4}
        &Avg. & 45.29 & \textbf{45.55}  \\
    \midrule
        \multirow{4}{*}{\begin{tabular}[c]{c@{}l@{}}Pair\\Classification\end{tabular}}&SDQ  & 93.07 & \textbf{93.32} \\
        &TSE  & 77.42 & \textbf{78.69} \\
        &TUC  & 87.18 & 86.73 \\
        \cmidrule(lr){2-4}
        &Avg. & 85.89 & \textbf{86.25}  \\
    
    \bottomrule
    \end{tabular}
    \end{center}

    \caption{
        Sentence representation performance on reranking, clustering, and pair classification tasks.
    }
    \label{tab:universal_gain}
    \vspace{-5pt}
\end{table}

In this section, we introduce details of our training settings, evaluation settings, main results, and further analysis of our method.
\subsection{Training Settings}
\paragraph{Training Dataset.}
We opt for the multi-task sentence-pair dataset MEDI~\cite{Su-2023-INSTRUCTOR}, comprising 330 sub-datasets spanning various tasks and domains, with a total of 1.4 million instances for training. Each sub-task is accompanied by corresponding natural language instructions elucidating its detailed goal or description.

\begin{table}[t]
    \begin{center}
    \centering
    \small
    \begin{tabular}{l|cccc|c}
    \toprule
       {Settings} & {BIO} & {S12}  & {S14} & {S22} & {Avg.}  \\
    
    \midrule
        
    \midrule
        \textbf{Data-CUBE} & \textbf{89.37} & \textbf{78.46}  & \textbf{83.06} & \textbf{68.28} & \textbf{84.41} \\  
        w/o Inst  & 87.53 & 77.64  & 82.85 & 66.95 & 83.94 \\
        w/o Task  & 86.56 & 78.15  & 82.53 & 67.48 & 83.94 \\
        Vanilla & 88.16 & 77.10  & 82.31 & 64.47 & 83.42 \\
    
    \bottomrule
    \end{tabular}
    \end{center}

    \caption{
        Ablations of the two-level curriculum.
    }
    \label{tab:abtest}
    \vspace{-5pt}
\end{table}

    
        
        
    


\begin{table}[t]
    \begin{center}
    \centering
    \small
    \begin{tabular}{l|cccc|c}
    \toprule
       {Settings} & {BIO} & {S12}  & {S14} & {S22} & {Avg.}  \\
    
    \midrule
        
    \midrule
    \textbf{Ours}   & \textbf{89.37} & \textbf{78.46}  & \textbf{83.06} & \textbf{68.28} & \textbf{84.41} \\
        800K  & 88.30 & 77.78 & 83.00 & 67.17 & 83.99 \\
        3M   & 88.40 & 77.71 & 83.06 & 68.12 & 84.10 \\
        5M   &  87.83 & 78.01 & 83.14 & 66.46 & 84.03 \\
    
    \bottomrule
    \end{tabular}
    \end{center}

    \caption{
        Variation studies of the iterations of Simulated Annealing algorithm on the test set of STS tasks.
    }
    \label{tab:abtest}
    \vspace{-5pt}
\end{table}

\begin{table}[t]
    \begin{center}
    \centering
    \small
    \begin{tabular}{c|cccc|c}
    \toprule
       {Settings} & {BIO} & {S12}  & {S14} & {S22} & {Avg.}  \\
    
    \midrule
        
    \midrule
        
        \textbf{Ours} & \textbf{89.37} & \textbf{78.46}  & \textbf{83.06} & \textbf{68.28} & \textbf{84.41} \\
        8 & 84.82 & 76.44 & 82.23 & 66.61 & 83.23 \\
        16 & 87.37 & 78.23 & 83.35 & 68.76 & 84.21 \\
        32 & 86.70 & 77.50 & 83.00 & 68.38 &  84.03 \\
    
    \bottomrule
    \end{tabular}
    \end{center}

    \caption{
        Variation studies of different batch sizes on the test set of STS tasks.
    }
    \label{tab:abtest}
    \vspace{-5pt}
\end{table}

\paragraph{Training Details}
To arrange the task-level curriculum, we utilize the simulated annealing algorithm and early stop at approximately 2 million steps to obtain a suboptimal task order $\mathcal{O}$. To rearrange the instance-level curriculum, we calculate the difficulty of instances~($\phi$ in Eq.~\ref{eq-ins_diff}) in advance and sort all the instances in each task by descending. After pre-reassigning the training data following Data-Cube, we start training from the checkpoint of INSTRUCTOR-large~(335M parameters)~\cite{Su-2023-INSTRUCTOR} with a batch size of 64. We use a softmax temperature $\tau$ of 0.01 and optimize the model with the AdamW optimizer. The warmup ratio is set to 0.1 and the learning rate is $2 \times 10^{-5}$.

\subsection{Evaluation Settings}
\paragraph{Evaluation Dataset}
We conduct evaluations on four task categories within the MTEB dataset~\cite{Muennighoff-2023-EACL-MTEB}, encompassing a total of 28 downstream tasks.
For STS tasks, we choose 10 datasets~(\eg STS12-27~\cite{Agirre-2012-STS12, Agirre-2013-STS13, Agirre-2014-STS14, Agirre-2015-STS15, Agirre-2016-STS16, Cer-2017-STS17}), to assess the performance. We employ Spearman's correlation of the English test set as the evaluation metric. Reranking tasks include AskUbuntuDupQuestions~\cite{Lei-2016-NAACL-askubuntu}, MindSmall~\cite{Wu-2020-ACL-Mind}, SciDocsRR~\cite{Cohan-2020-ACL-SciDocsRR}, and StackOverflowDupQuestions~\cite{Liu-2018-SIGSOFT-StackOverflowDupQuestions}. Mean Average Precision (MAP) of the test set is utilized to measure performance.
Clustering tasks encompass 11 datasets, such as MedrxivClusteringS2S~\cite{Muennighoff-2023-EACL-MTEB} and StackExchangeClusteringP2P~\cite{Geigle-2021-arxiv-RedditClustering}.
In these clustering tasks, we use v-measure of the test set as the evaluation metric. Pair Classification tasks consist of SprintDuplicateQuestions~\cite{J.Shah-2018-EMNLP-Adversarial}, TwitterSemEval2015~\cite{Xu-2015-ACL-semeval2015}, and TwitterURLCorpus~\cite{Lan-2017-EMNLP-Lan}. Performance is assessed using accuracy on the test set.

\paragraph{Baseline Models}
We select several sentence representation methods that have achieved state-of-the-art performance on STS tasks, including publicly available models scaling from 100M to 11B parameters, and APIs without exact parameter amounts.
Concretely, we select traditional no-instruction sentence representation models~(\eg GloVe~\cite{Pennington-2014-EMNLP-glove}, USE~\cite{Cer-2018-EMNLP-USE}, Contriever~\cite{Izacard-2022-TMLR-Contriever}, GTR~\cite{Ni-2022-GTR}, SimCSE~\cite{Gao-2021-SimCSE}, SGPT~\cite{Muennighoff-2022-SGPT}, E5~\cite{Wang-2022-arxiv-E5}, and SentenceT5~\cite{Ni-2022-ACL-SentenceT5}), instruction-based models~(\eg Jina~\cite{Gunther-2023-arxiv-Jina}, Udever~\cite{Zhang-2023-arxiv-Udever}, Stella~\footnote{https://huggingface.co/infgrad/stella-base-en-v2}, BGE~\cite{Xiao-2023-BGE}, GTE~\cite{Li-2023-GTE}, and INSTRUCTOR), and APIs~(\eg OpenAI Text Embedding~\cite{Neelakantan-2022-Text}, Voyage~\footnote{https://docs.voyageai.com/}, Cohere~\footnote{https://txt.cohere.com/introducing-embed-v3/}, and Ember~\footnote{https://docs.llmrails.com/embedding/embed-text}).


\subsection{Main Results}
We present the main experiment results in Table~\ref{tab:main_sts}. 
Based on the results, it is evident that instruction-based sentence representation models generally perform better than no-instruction models, although some no-instruction models are much larger than instruction-based models in terms of scale~(\eg Sentence-T5 XXL has 11B parameters while the largest version of BGE is about 300M). A potential reason is that instruction tuning enhances the models' understanding of tasks by integrating natural language instructions. This integration assists the models in encoding sentences into task-aware representations, thereby providing a significant advantage in downstream tasks.

Among all the compared models, our method achieves the highest average performance across STS tasks. Although it may not always rank first in certain tasks like STS-12, STS-13, STS-16, STS-22, and STSBenchmark, it maintains competitive results.
Notably, models achieving the best results in these tasks tend to excel in only one task but underperform in others. In contrast, our approach consistently demonstrates effectiveness across all tasks. When compared to our backbone model~(\ie INSTRUCTOR), our method significantly improves the performance on all STS tasks. 
In addition, we also evaluate our method on other task categories~(\eg  Reranking, Clustering, and PairClassifiacation). As Table~\ref{tab:universal_gain} shows, our method achieves average performance gains on these diverse task categories.
This implies that our approach plays a significant role in reducing the interference among tasks, which enhances not only the performance in specific tasks but also the overall versatility.

It is noteworthy that both the volume of training data and the mini-batch size utilized in our approach are considerably smaller compared to other robust instruction-based sentence representation models. Specifically, our model is trained with only 1 million sentence pairs and a batch size of 64, in stark contrast to models like BGE which use 300 million sentence pairs and a batch size of 32768.
This indicates that our approach has the potential to enhance the model's ability to learn more effectively from multi-task data in the context of data interference and achieve comparable or even superior performance despite limited data and computational resources.

\subsection{Further Analysis}
Next, we continue to investigate the effectiveness and robustness of Data-Cube. This involves conducting ablation studies on the two-level curriculums, closely assessing the impact of iterations in Simulated Annealing within the task-level curriculum, analyzing the influence of mini-batch size during training, and thoroughly examining the training convergence. To gain a better understanding of the variations between different settings, we carefully select several tasks in STS, such as BIOSSES~\cite{Sogancioglu-2017-BIOSSES}, STS-12~\cite{Agirre-2012-STS12}, STS-14~\cite{Agirre-2014-STS14}, and STS-22~\cite{Chen-2022-STS22}, where our method demonstrates more noticeable improvements.

\subsubsection{Ablations of Two-level Curriculum}  To explore the influence of task-level curriculum, we exclusively implement the instance-level curriculum to reorganize the training data. This entails arranging instances within each task based on their difficulty using Eq.~\ref{eq-ins_diff}, while the task orders are randomly shuffled. Conversely, we employ the task orders generated by the task-level curriculum but randomly shuffle the instances within each task to assess the effectiveness of the instance-level curriculum. Furthermore, we proceed with the training of INSTRUCTOR without any additional operations, treating this as the vanilla performance baseline. As Table~\ref{tab:abtest} shows, both task-level curriculum and instance-level curriculum contribute to alleviating the interference of multi-task data and improving the performance of the sentence representation model.



\subsubsection{Iterations of Simulated Annealing} In the task-level curriculum, we employ Simulated Annealing algorithm that gradually obtains an approximate solution as the iterations progress. 
In broad terms, a higher number of iterations typically leads to a more optimal solution. Consequently, we undertook experiments on task orders generated through varying iteration counts to illustrate the adequacy of the specific iteration count we employed. We opt to compare the results of using task orders generated by SA at 800K, 2M~(Ours), 3M, and 5M iterations. 
As Table~\ref{tab:abtest} shows, these performances are comparable, indicating that the chosen iteration step of 2M is adequate for alleviating interference across tasks. 

\subsubsection{Size of Mini-batch} During multi-task contrastive learning, we leverage the in-batch negatives to extend the positive-negative ratio. This approach has been shown to enhance the uniformity of the sentence representation model and thereby improve overall performance~\cite{Karpukhin-2020-DPR}
In the instance-level curriculum, we propose to alleviate the interference between instances with different difficulty. However, when using a larger mini-batch, the variability in difficulty within the batch increases, seemingly contradicting our curriculum design. To address this concern, we conduct experiments to assess the extent to which the instance-level curriculum enhances model performance under different mini-batch sizes.

As depicted in Table~\ref{tab:abtest}, while a larger mini-batch size contributes to superior overall performance, the results with Data-CUBE remain comparable even in a smaller batch size. Intriguingly, the model demonstrates better performance with a batch size of 16 compared to 32.  This suggests that our approach is particularly beneficial in scenarios with limited hardcore resources.


\begin{figure}[t]
    \centering
    \includegraphics[width=0.46\textwidth, trim=25 10 10 25]{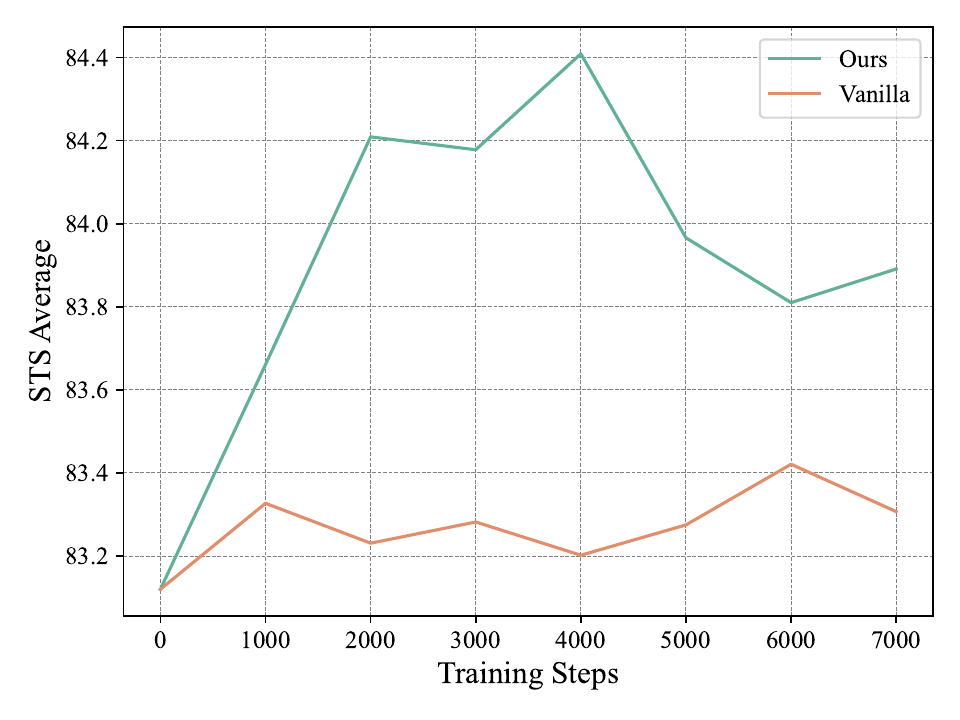}
    \caption{Performance variation curve on the STS tasks during the training process. }
    \label{fig:train_convergence}
\end{figure}

\subsubsection{Analysis of Training Convergence}
To validate the efficacy of our approach in mitigating data interference, we assess the model's performance on STS tasks throughout the training process~(See Figure~\ref{fig:train_convergence}). In comparison to directly continuing training, employing our curriculum leads to significantly improved performance in fewer training steps. Furthermore, we compare the ratio of underfitting instances within tasks before and after training with Data-CUBE~(See Figure~\ref{fig:underfitting_degree}). It is evident that the ratio of underfitting instances consistently decreases across various tasks.  This suggests that our method effectively assists the model in alleviating interference in multi-task contrastive learning and results in a more powerful and robust sentence representation model.

%% file: sec-con.tex
\section{Conclusion}
In this work, we proposed Data-CUBE, a data curriculum method for multi-task instruction-based sentence representation learning.
Our core idea is to reduce the cross-task and cross-instance interference risks caused by the randomly ordered tasks and sampled instances with large divergence, through arranging their orders before training.
To achieve this, we employed a simulated annealing algorithm to find the optimal task order to minimize the cross-task interference, and assigned all instances per task into easy-to-difficult mini-batches to reduce the cross-instance interference.
Experimental results on MTEB sentence representation evaluation tasks have shown that our approach can boost the performance of state-of-the-art baselines.

In the future, we will apply our method for other tasks and the pre-training process of large language models.
Besides, we will also explore a more efficient and effective data curriculum method for large-scale sentence representation learning.